# The Univariate Flagging Algorithm (UFA): a Fully-Automated Approach for Identifying Optimal Thresholds in Data

Mallory Sheth[1,2], Roy Welsch[1], Natasha Markuzon[2]*

[1] Massachusetts Institute of Technology, Cambridge, MA, USA
[2] The Charles Stark Draper Laboratory, Cambridge, MA, USA
*Corresponding author, Email: nmarkuzon@draper.com

## Abstract

In many data classification problems, there is no linear relationship between an explanatory and the dependent variables. Instead, there may be ranges of the input variable for which the observed outcome is signficantly more or less likely. This paper describes an algorithm for automatic detection of such thresholds, called the Univariate Flagging Algorithm (UFA). The algorithm searches for a separation that optimizes the difference between separated areas while providing the maximum support. We evaluate its performance using three examples and demonstrate that thresholds identified by the algorithm align well with visual inspection and subject matter expertise. We also introduce two classification approaches that use UFA and show that the performance attained on unseen test data is equal to or better than that of more traditional classifiers. We demonstrate that the proposed algorithm is robust against missing data and noise, is scalable, and is easy to interpret and visualize. It is also well suited for problems where incidence of the target is low.

# Introduction

In many data classification problems, there is no linear relationship between an explanatory and the dependent variables. Instead, there may be ranges of the input variable for which the observed outcome is signficantly more or less likely. In clinical decision making, for example, doctors identify ranges of laboratory tests values that may identify patients' higher risk of developing or having a disease [1, 2]. In earth science, amount of rainfall thresholds can be used to develop early warning systems for landslides or flooding [3, 4].

Many nonlinear classifiers, such as decision trees [5] or support vector machines (SVM) [6], are designed to find "optimal" cutpoints, typically defined as cutpoints that minimize some measure of node impurity. Such measures include misclassification rate, Gini index, or entropy/information gain [5, 6]. Supervised clustering works similarly, minimizing impurity while adding a penalty for the total number of clusters [7]. Alternatively, Williams, et al (2006) put forth a minimum p-value approach for finding optimal cutpoints for binary classification. Their algorithm uses a chi-squared test to find the cutpoint that maximizes the difference in outcomes between both sides [2].

These approaches are similar in that they consider the entire input space, both false positives and false negatives, to select the optimal cutpoint. In certain applications, however, one may care more about a subspace of increased incidence of a target. Under certain conditions, it might be important to identify separation thresholds that are associated with a high prevelance of the target, while the overall solution is not optimized. Examples include medical conditions where values outside clinically defined thresholds are associated with high mortality, while more normal values may not provide much information.

For example, in individuals with a condition called sepsis, low body temperature is associated with illness severity and death [8]. Figure 1 displays average body temperature for 512 septic patients, with an overall death rate of 30.9%. Patients who died are denoted in red, while patients who survived are denoted in blue. International guidelines for sepsis management define low body tempeature as 36° C [9]. For patients below this threshold death rate stands at 57.1%, nearly twice the overall death rate, while little can be said about patients above the threshold (Figure 1).

**Figure 1: Body temperature for adult sepsis patients**

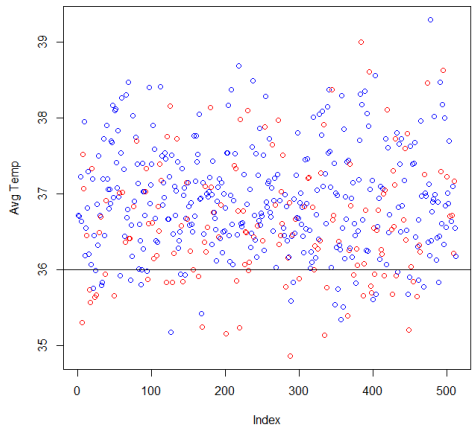

We propose an algorithm for identifying such thresholds in an automated fashion. In the decision tree or SVM framework, cost functions penalizing for false positives or negatives will shift the "optimal" cutoff to satisfy the request [5, 6]. In practice, however, it is often difficult to quantify the costs associated with different types of errors, in particular in the medical domain.

Friedman & Fisher's (1999) Patient Rule Induction Method (PRIM) procedure finds rectangular subregions of the feature space that are associated with a high (or low) likelihood of the outcome.The subregions are then slowly made smaller, each time increasing (or decreasing) the rate of the outcome [10]. With this method and others like it, there is an inherent trade off between the number of data points within the subregion (the support) and the proportion of the datapoints that are associated with the outcome (purity), where smaller supports generally have higher purity. With PRIM, the user is responsible for defining the "optimal" subregion, by specifying the prefered trade off for the application. While this may work well in some situations, identifying the appropriate trade off is challenging, suggesting the need for an algorithm that requires less user input.

In this paper, we put forth a threshold detection algorithm called the Univariate Flagging Algorithm (UFA). UFA optimizes over subregions of the input space, but performs the trade off between support and purity automatically. We show that UFA can identify the existence of thresholds for individual variables and that they align with visual inspection and thresholds established by subject matter experts. We also demonstrate that these thresholds can be used to classify previously unseen test cases with performance equal to or better than many commonly used classification techniques, such as the random forest and logistic regression. Moreover, the UFA system easily scales to a large number of variables, is robust against missing data and noise, is capable of predicting rare events, and is easy to interpret and visualize, making it appealing for a wide range of real-world applications.

## Methods

UFA is designed to identify an optimal cutpoint for a single explanatory variable that is associated with a significantly higher or lower likelihood of the target. UFA identifies up to two thesholds, one below the median and one above the median. The algorithm is intended for a binary target $y$ (e.g. [0, 1]) and a continuous explanatory variable $x$. At its most basic level, UFA finds the value $x = x^{opt}$ that maximizes the difference in the outcome rate for observations that fall outside $x^{opt}$ and a baseline rate, while maintaining a good level of support.

*Formal specification*

The following variables are necessary for the formal specification of the UFA algorithm (Table 1). For the purpose of formulation, we consider candidate thresholds below the median value of $x$.

**Table 1: List of variables for specification of UFA algorithm.** For the purpose of formulation, we consider candidate thresholds below the median value of $x$.

| Variable | Definition |
| --- | --- |
| $y \in \{0,1\}$ | Binary target |
| $x \in [x_{min}, x_{max}]$ | Continuous explanatory variable |
| $x_{iqr} \in [x_{P25}, x_{P75}]$ | Values of $x$ in the interquartile range, defined as 25th to 75th percentile |
| $n_{iqr}$ | Number of observations in $x_{iqr}$ |
| $p_{iqr}$ | Percent of $n_{iqr}$ with $y = 1$ |
| $\hat{x}_i$ | Candidate threshold below median of $x$ |
| $n_{i-}$ | Number of observations below candidate threshold |
| $p_{i-}$ | Percent of $n_{i-}$ with $y = 1$ |

For each $\hat{x}_i$, we conduct the following hypothesis test to check for a significant difference in the outcome rate below the threshold and the outcome rate in the interquartile range:

$$H_0: \quad p_{i-} - p_{iqr} = 0 \qquad (1)$$

$$H_a: \quad p_{i-} - p_{iqr} \neq 0$$

We are using a binomial proportion test [11] with test statistic $Z_i$:

$$Z_i = \frac{p_{i-} - p_{iqr}}{\sqrt{p_{i-}^{wa} * (1 - p_{i-}^{wa}) * (\frac{1}{n_{iqr}} + \frac{1}{n_{i-}})}} \qquad (2)$$

where $p_{i-}^{wa}$ is the weighted average of the outcome rates, calculated:

$$p_{i-}^{wa} = \frac{(p_{iqr} * n_{iqr} + p_{i-} * n_{i-})}{n_{iqr} + n_{i-}} \qquad (3)$$

We define $\hat{x}^{opt}$ as the candidate threshold $\hat{x}_i$ with the maximum $Z_i$ in absolute value:

$$\hat{x}^{opt} = \max_{\hat{x}_i}[abs(Z_i)] \qquad (4)$$

$Z_i$ provides an inherent trade-off between maximizing the support and maximizing (or equivalently, minimizing) the outcome rate. The proposed measure does not provide an optimal separation in terms of minimizing the overall misclassification rate, but is optimized against finding areas enriched with cases with target outcome. The same applies to finding areas with specifically low rate of the target.

*Procedure to find optimal threshold for variable $x$:*

1. Generate a list of potential thresholds $\hat{x}_{i-}$ between the median value of $x$ and the minimum value of $x$, excluding those with low support, by dividing the range into $n$ segments.
   - For the purpose of this paper, we excluded the five lowest values of $x$, assuming that thresholds with a support fewer than five are of no interest.
   - Currently, we consider 50 segments of equal length.

2. Calculate $Z_i$ as specified in equation (2). Define $\hat{x}^{opt}$ according to equation (4).

3. Check $\hat{x}^{opt}$ for statistical significance by comparing its Z-value to a chosen critical value. Keep the threshold if it is significant and discard it otherwise.
   - For the purpose of this paper, we used a critical value of 2.576 to establish significance, which is associated with a p-value of 0.01. We address issues related to multiple testing by validating the thresholds on previously unseen data.

Through this procedure, UFA finds the optimal threshold below the median for each variable $x$. The procedure can then be repeated for area above the median.

*Classification using UFA*

UFA is designed to work with a single variable. We used its results as features for multidimensional classifiers. There are many approaches to incorporate UFA-designed thresholds into a multi-dimensional classifier, we present two possibilities in this paper. Both create an indicator variable or "flag" for each signficant threshold, which takes the value of one if the value of the variable in observation exceeds the threshold and zero otherwise.

**Number of Flags algorithm (N-UFA).** The first classifier aggregates the number of "high risk" and "low risk" flags for each observation, creating a two-dimensional vector for each observation. Then, a linear decision boundary is drawn to separate one class from the other along these two dimensions. Throughout the paper, this approach will be denoted as the Number of Flags algorithm (N-UFA). Figure 2 shows an example of the N-UFA classifier's performance in predicting adult sepsis patients' mortality. For each patient, we count the number of flags that are associated with a high likelihood of mortality and the number of flags that are associated with a low likelihood of mortality; the solid line represents the linear decision boundary that minimizes the misclassification rate along these two-dimensions. Throughout this paper, each flag receives an equal weight of one, though future research could investigate the impact of assigning flags different weights.

**Figure 2: Number of high mortality and low mortality flags for adult sepsis patients.** Patients who died are indicated by red squares while patients who lived are indicated by blue triangles. For each patient, we counted the number of flags that are associated with a high likelihood of mortality and the number of flags that are associated with a low likelihood of mortality; the solid line represents the linear decision boundary that minimizes the misclassification rate along these two dimensions.

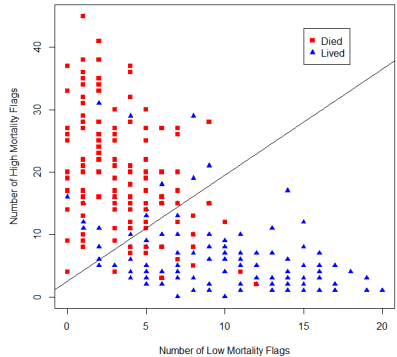

**UFA-created thresholds as features in Random Forest (RF-UFA)**. We used UFA-identified flags as independent dummy features in a random forest model [5]. We will show below that under certain conditions the classification performance is comparable or better than using original features.

Throughout the results section, we compare the accuracy and area under the receiver operating curve (AUROC) of the two UFA-based classifiers to other commonly used classification techniques. We also highlight a number of practical advantages of the UFA system in general and N-UFA in particular, including its interpretability, scalability, ability to handle missing and noisy data, and ability to predict rare events.

## Results

In this section, we apply the UFA system to three different datasets, the Iris dataset, MIMIC II clinical dataset, and Seattle landslide dataset that are discussed in more detail below. These datasets differ significantly in data complexity and target/non-target ratio. We demonstarte that for all three datasets the UFA systems performs equally or better than commonly used classifiers when evaluated on previously unseen data.

We demonstrate that the thresholds automatically generated by the UFA algorithm align well with visual inspection and subject matter expertise, and that the results are achieved with little or no a priori knowledge of the data. We also demonstrate practical advantages of the UFA system, such as its scalability, ability to handle missing and noisy data, its ability to predict rare events, and the fact that it greatly reduces problem dimensionality enabling easy interpretation and visualization.

*Data Description*

The first application is the well-known Iris dataset [12]. A classic in the machine learning discipline, it contains 50 observations for three different species of Iris. One of the species, *Iris setosa*, is linearly separable while the other two species, *Iris virginica* and *Iris versicolor*, are not. We ran UFA over this relatively straighforward dataset to ensure that it performed comparably to other standard approaches.

The second application is substantially more complex. The publicly available MIMIC II database, version 2.6, contains de-identified clinical data for over 30,000 adult intensive-care unit (ICU) stays [13]. Focusing on patients admitted with a primary diagnosis of sepsis, we processed over 200 variables covering the first four days of the patient's stay. These variables included both static features like demographics, as well as dynamic features such as trends in laboratory values or vital signs. As outlined in the introduction, our final dataset contained 512 patients with a mortality rate of 30.9%.

The final dataset that we used to evaluate UFA contains all reported landslides for Seattle, WA from 1965 to 1999 [14]. Each observation in the dataset represents one day, and contains information on precipitation, temperature, and wind, along with whether a landslide was reported. Of the nearly 13,000 days in the dataset, only 2.3% had one or more landslide. This dataset was included to evaluate UFA's ability to predict rare events.

*Thresholds identified with UFA align with visual inspection and subject matter expertise*

One way to evaluate UFA's performance is to evaluate the validity of the thresholds. In some cases, the optimal cutpoint is clear, such as the trivial case when the data is linearly separable. In other instances, thresholds may represent a known physical or biological boundary; in these situations, we can compare our thresholds to the known values.

We use the Iris data to illustrate the trivial case. In Figure 3, it is clear that *Iris setosa,* denoted in red, is linearly separable from the other two classes, which are denoted in blue. As expected, UFA is able to successfully identify a threshold for a single variable, 'petal length', which separates the two classes from one another.

**Figure 3: Petal lengths for different species of *Iris*.** *Iris setosa,* denoted in red, is linearly separable from the remaining instances, which are denoted in blue. UFA is able to successfully identify a threshold for a single variable which separates the two classes from one another.

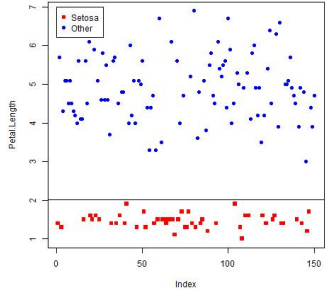

For a more complicated example, we return to Figure 1 from the intoduction of the paper. That figure displays body temperature for patients with sepsis, and includes a cut-point at 36° C aligning with the clinical definition of low body temperature. In sepsis, low body temperature is one of the diagnostic criteria for severe sepsis and septic shock, and is known to be associated with patient severity and death [8]. Applying UFA to the MIMIC II data for body temperature, we identify a high-mortality threshold at 35.97°C which aligns closely with the known physiological limit. Below the threshold of 35.97°C, sepsis patients die at a rate of 57.9%, nearly twice the overall death rate.

Table 2 shows three other examples of variables in the MIMIC II database with known clinical thresholds. For each variable, UFA identifies a significant threshold that is well within one standard error of a known bound, as established by the National Institutes of Health [15]. The mortality rates for patients who violated these thresholds range from 52.7% to 55.9%, much higher than the 30.9% death rate in the septic population overall. Moreover, though UFA only identified one significant threshold for each variable in Table 1, the directionality is consistent with clinical understanding of sepsis. For example, it is well-known that hypotension, or *lowered* blood pressure, is associated with worsening sepsis [9], which is consistent with our findings.

**Table 2: Examples of UFA-defined thresholds, MIMIC II data.** For each variable in the table, the UFA-identified threshold aligns with the known physiological bound, as established by the National Institutes of Health. The mortality rates for patients who violated these thresholds range from 52.7% to 55.9%, much higher than the 30.9% death rate in the septic population overall.

| Variable | Normal Range | Threshold | | Support | Mortality |
|---|---|---|---|---|---|
| Phosphorus Level | 2.4 - 4.1 mg/dL | More Than | 4.5 | 93 | 52.7% |
| Sodium Level | 135 - 145 mEq/L | Less Than | 134.9 | 59 | 55.9% |
| Mean Arterial BP | 70 - 110 mmHg | Less Than | 67.4 | 86 | 55.8% |

*UFA-based classifiers have predictive performance equal to or better than existing methods*

A second way to evaluate UFA's performance is to determine whether it can correctly predict the class of previously unseen data. Once again, we begin with the Iris dataset as it is our most straightforward application. The Iris dataset contains three classes. However, in the previous section, we showed that UFA can linearly separate *Iris setosa* from the other two. Therefore, in this section, we primarily focus on identifying thresholds that separate the two remaining classes, *Iris versicolor* and *Iris virginica*.

The Iris dataset contains four variables and, therefore, UFA searches for up to eight possible thresholds in the data. Table A1 in the supplemental materials contains the optimal thresholds for each variable and shows that six of the eight are significant. In Table A1, the automatic trade-off between purity and support inherent to UFA is apparent. While a sepal width less than 2.4 identifies a subset of cases where 90% belong to the class *versicolor*, the support (N=10) is not large enough to consider this variable in subsequent analysis. The other variables, however, each have two signficant thresholds. Visualizations are also available in the supplemental materials.

We convert the significant thresholds into indicator variables ("flags") which take the value one if the data point falls outside the threshold and zero otherwise. Plugging these flags into RF-UFA, we find that we can correctly classify 48 of 50 cases for both *Iris versicolor* and *Iris virginica*. We achieve the same level of accuracy using N-UFA. This performance is in line with the apparent error rate of other classification algorithms that have been used on the Iris dataset [16, 17].

We can also use these approaches to predict the class of previously unseen cases. Using five-fold cross validation, both UFA-based classifiers have 100% accuracy separating *Iris setosa* from the other classes. Overall, N-UFA averages 94.7% accuracy across the five folds, while RF-UFA achieves an average accuracy of 96.0%. Once again, this performance is consistent with other commonly used classifiers.

Next, we apply UFA to mortality prediction in sepsis patients. Running UFA for all 218 variables created using MIMIC II dataset, we identified 95 thresholds associated with high mortality and 43 thresholds associated with low mortality in sepsis patients. Figure 2 (from the Methods section) show a plot for patients according to their number of high and low flags, where red denotes patients who died and blue denotes patients that lived. A linear decision boundary effectively separates the two classes.

Using ten-fold cross validation, on average, N-UFA correctly classifies 77.5% of test cases, while RF-UFA acheives 78.1% accuracy. As seen in Table 3, this performance is better than or comparable to classifying patients based on the original, continuous data for a variety of commonly used linear and non-linear methods. Similarly, the AUROC for the two UFA-based classifiers is signficantly higher than all of the non-UFA methods with the exception of random forest.

**Table 3: Comparison of different classifiers for in-hospital mortality of adult sepsis patients.** The two UFA-based classifiers have predictive performance better than or equal to other commonly used classification techniques.

| Classifier | | Accuracy | AUROC |
|---|---|---|---|
| N-UFA | UFA-based | **77.5%** (75.1, 79.9) | **0.819** (0.797, 0.841) |
| RF-UFA | | **78.1%** (75.8, 80.3) | **0.800** (0.779, 0.821) |
| Logistic Regression | Other | 68.7% (65.7, 71.6) | 0.698 (0.642, 0.753) |
| Support Vector Machine | | 79.4% (76.2, 82.6) | 0.555 (0.331, 0.780) |
| Decision Tree | | 68.8% (66.0, 71.7) | 0.626 (0.575, 0.677) |
| Random Forest | | 79.0% (76.9, 81.1) | 0.823 (0.796, 0.851) |

In our last application, we use UFA to predict the rare event of landslides. Of the nealy 13,000 days described in the Seattle database [14], only 2.3% had one or more landslide. Given the relative infrequency of the target, we focused our analysis on identifying variables or groups of variables that were predictive of an increased risk.

UFA identified 32 signficant thresholds associated with an increased likelihood of landslide. For example, when precipitation for the last four days exceeds 3.2 inches, the percentage of days with a landslide is 34.3%, nearly 15x the rate for a typical day. Moreover, as seen in Figure 4, thresholds can be combined to find conditions under which the relative risk of a landslide is even higher. If one combines the rain threshold with a maximum daily wind of more than 7.8, the percentage of days with a landslide jumps to 51.4%, more than 22x the rate for a typical day.

**Figure 4: Landslide days stratified by precepitation and wind.** The percentage of days with a landslide in each quadrant is displayed in red. Thresholds can be combined to find conditions under which the relative risk of a landslide is signficantly elevated.

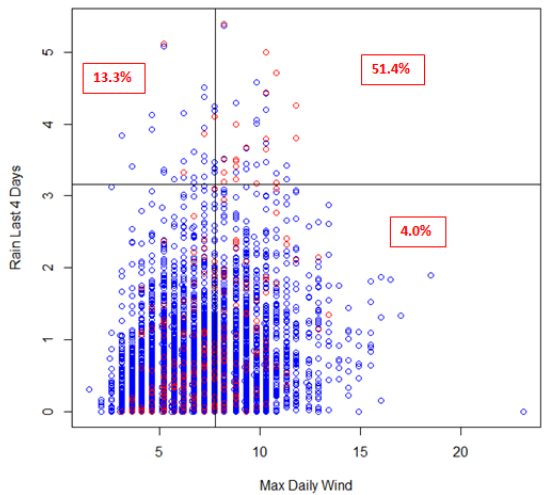

In general, it is difficult to train classifiers when the incidence of the target is very low [18] . This is because the classifier can achieve very high accuracy by always predicting the more likely outcome; in this case, a model that always predicts no landslide would have accuracy of 97.7%. One solution is to balance the training dataset, so that it has an equal number of days with and without a landslide. However, this may be undesirable for a variety of reasons. In particular, for very rare events, balancing the dataset through undersampling may exclude a large number of potentially useful majority-class examples, while balancing the dataset through oversampling can lead to overfitting [19].

We find that one advantage of N-UFA is that it can identify days that are at high risk of landslide, even with unbalanced training data. Using 80% of our landslide dataset for training, we find that days with 12 or more flags are 14.8x more likely to have a landslide. If we consider this the definition of a "high risk" day, and apply the same criteria to the remaining 20% of the data, we see that this definition generalizes well. The rate of landslides on high risk days in the test set is 18.2%, almost 8x the typical rate.

*The UFA-based number of flags classifier is robust to noise and missing data*

Because each variable is considered individually in UFA, there is no need to have complete data for each observation. If an instance is missing data for a particular variable, it can simply be excluded from the calculation of that variable's threshold, but remain included in calculations for which data is present.

The question remains, however, whether the UFA-based classifiers will have high predictive power if certain flags are missing or assigned incorrectly due to noisy data. We hypothesize that N-UFA in particular should be robust to noise and missing data, since it aggregates over all of the high and low risk flags and does not depend on individual variables.

Table 4 confirms this hypothesis using the MIMIC II data. It compares the performance of N-UFA, random forest, and logistic regression for the original MIMIC II data and a version of the MIMIC II data where 50% of observations were replaced randomly with missing values. As discussed earlier, N-UFA has the advantage of not requiring complete data for each observation, so imputation was perfromed for it. However, logistic regression and random forest do not share this characteristic. To avoid excluding patients with missing data, we imputed missing values for these methods using the sample average[1].

**Table 4: Comparison of different classifiers with varying amounts of missing data.** This table compares the performance of different classifiers for the original MIMIC II data and a version of the MIMIC II data where 50% of observations were replaced randomly with missing values. We see that N-UFA is robust to missing data, with accuracy decreasing just 1.3% as the amoung of missing data increases to 50%. An expanded versions of Table 4 including confidence intervals and results for 5-25% missing data is available in the supplemental materials.

| Classifier | | Accuracy | | | AUROC | | |
|---|---|---|---|---|---|---|---|
| | | 0% | 50% | Δ | 0% | 50% | Δ |
| N-UFA | UFA-Based | 77.5% | 76.2% | **1.3%** | 0.819 | 0.790 | **0.029** |
| Random Forest | Other | 79.0% | 71.9% | **7.1%** | 0.823 | 0.771 | **0.052** |
| Logistic Regression | | 68.7% | 58.3% | **10.4%** | 0.698 | 0.598 | **0.100** |

Random forest was included in Table 4 because it was the non-UFA based algorithm with the highest accuracy and AUROC under 50% missing data. Logistic regression was included as an additional comparison. Table 4 shows that with 50% missing data, N-UFA has the highest accuracy and AUROC of all three methods. We also see that the difference in accuracy between 0% missing data and 50% missing data for the N-UFA approach is only 1.3 percentage points, compared to 7.1 percentage points for random forest and 10.4 percentage points for logistic regression. Similarly, AUROC decreases by 0.029 as opposed to 0.052 and 0.100 respectively.

Table 5 provides similar results for data accuracy. It presents accuracy and AUROC for N-UFA, random forest, and logistic regression when 50% of the MIMIC II data is randomly perterbed by a value $\epsilon$, distributed normally with mean zero and the empirical variance of the variable in question. Once again, we see that N-UFA holds up well. With 50% imprecise data, the accuracy and AUROC are in line with random forest and signficantly higher than logistic regression. On average, the accuracy decreases by just 1.7 percentage points for N-UFA, while AUROC decreases by 0.230 as the percentage of imprecise data increases to 50%.

---

[1] While other imputation approaches exist, a full survey is outside the scope of this paper.

**Table 5: Comparison of different classifiers with varying amounts of imprecise data.** This table compares the performance of different classifiers for the original MIMIC II data and a version of the MIMIC II data where 50% of observations were randomly perterbed by a value $\epsilon$, distributed normally with mean zero and the empirical variance of the variable in question. We see that N-UFA is robust to imprecise data, with accuracy decreasing just 1.7% as the amoung of imprecise data increases to 50%. An expanded versions of Table 5 including confidence intervals and results for 5-25% imprecise data is available in the supplemental materials.

| Classifier | | Accuracy | | | AUROC | | |
|---|---|---|---|---|---|---|---|
| | | 0% | 50% | Δ | 0% | 50% | Δ |
| N-UFA | UFA-Based | 77.5% | 75.8% | **1.7%** | 0.819 | 0.796 | **0.023** |
| Random Forest | Other | 79.0% | 76.3% | **2.7%** | 0.823 | 0.802 | **0.021** |
| Logistic Regression | | 68.7% | 68.8% | **-0.1%** | 0.698 | 0.681 | **0.017** |

Expanded versions of Table 4 and Table 5 including confidence intervals and results for 5-25% missing or imprecise data are available in the supplemental materials. These additional results support the conclusion that the UFA system is robust to missing and noisy data, with performance equal to or exceeding random forest and logistic regression.

*The UFA system is robust to small variations in the risk thresholds*

The UFA system selects a single 'optimal' threshold where optimality is defined as the maximum absolute Z-statistic in the training dataset. However, a different training dataset may produce a different optimal threshold (which may or may not be significant). In this section, we show that the UFA system is robust to variations in the thresholds that arise through changes to the training data.

To do this, we employ bootstrapping, a data-driven technique where one resamples the training data many times to generate additional new training datasets. We ran UFA on each new dataset to create a histogram of possible thresholds for each variable. Returning to the MIMIC II database, Figure 5 shows 1,000 bootstrapped thresholds for body temperature. The vertical line at 35.97°C represents the optimal cut point $\hat{x}^{opt}$ that was found using the full training data.

**Figure 5: Bootstrapped thresholds for low body temperature in adult sepsis patients.** 1,000 bootstrapped thresholds for body temperature. The vertical line at 35.97°C represents the optimal cut point that was found using the original training data. The figure shows that this value aligns with the mode of the distribution of bootstrapped thresholds, providing support for its validity.

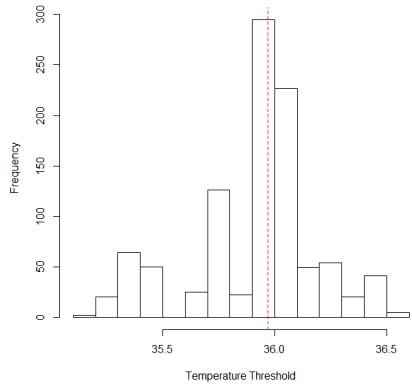

The information in this histogram is useful in two ways. First, we can calculate the variance in the potential thresholds, which can help quantify uncertainty. Second, we can compare $\hat{x}^{opt}$ for the full training data to the bootstrapped distribution, and determine whether it is consistent with the other trials. If we are concerned that $\hat{x}^{opt}$ may be overfit to the training data and not generalizable, we can consider using a feature of the bootstrapped distribution such as the mean or mode as our candidate threshold instead. For the sepsis application, we ran 100 bootstraps per variable and used the mode of each distribution as the candidate threshold, instead of $\hat{x}^{opt}$. On average, we found 5% fewer signficant thresholds and, of the significant thresholds found, 15.6% varied by more than 5% from $\hat{x}^{opt}$.. When applied to unseen data, however, N-UFA achieved the exact same AUROC using the bootstrapped thresholds as it did using the original thresholds suggesting that N-UFA is robust to small variations in the thresholds.

Since bootstrapping adds significant runtime to the UFA system and did not improve the predictive performance for our application, we present the non-bootstrapped results in this paper. However, this is a possible area for further work. In particular, visual inspection of the boostrapped distributions for the MIMIC II data reveals that some of the variable have a bimodal distribution of boostrapped thresholds, perhaps suggesting multiple cut points.

## Discussion

UFA automatically detects target-defined separation thresholds in data. In the results section, we showed that the thresholds that it detects align with subject matter expertise and can be used to classify previously unseen data with performance equal to or better than other standard classifiers. We also showed that the UFA system has a number of desirable characteristics that make it well-suited for general use, such as scalability, interpretability, ability to handle missing and noisy data, and ability to predict rare events.

The UFA system is designed to handle the challenges of big data. Since UFA runs on each variable individually, it can easily be applied to datasets with a very large number of features, including cases when the number of features is much larger than the number of observations. Thresholds for each variable can be identified in parallel, allowing for efficient computing. Further, though UFA is univariate, the ability to quickly and automatically consider a large number of features means that researchers can easily introduce new variables that are interactions of existing features, if thought to be important for the application.

As data becomes larger, one common difficulty is an increase in the number of missing values. This is particularly problematic for classifiers that drop observations that do not have complete data, since this can significantly decrease the number of observations available for analysis. The UFA system has the advantage of not requiring complete data; if an instance is missing data for a particular variable, it can simply be excluded from the calculation of that variable's theshold, but remain included in calculations for which data are present. In the results section, we showed that the UFA system holds up well to large amounts of missing data, as well as imprecise or noisy data which can be common in many real-world applications.

Another challenge of big data is interpretability. As the number of dimensions become large, it can become difficult to visualize a classifier's decision criteria and to understand the relationship between individual variables and the outcome. UFA, however, gives the user a list of the variables with significant thresholds, along with the difference in the outcome rates and the support.  The thresholds themselves are easy to understand, interpret, and verify against pre-existing domain knowledge. Classification is straightforward as well. N-UFA in particular has the advantage of being two-dimensional, making it easy to visualize. One can simply create a plot with high risk flags on one axis and low risk flags on the other, along with the relevent decision boundary (example in Figure 2). New cases can be added, and the user can easily see where the instance falls both in terms of its classification, as well as its distance from the decision boundary.

Formal specification of conditions under which the UFA system breaks down was outside the scope of this paper, and is an area for future research. However, the three applications in this paper demonstrate that it works well in a variety of different scenarios.  The three example datasets vary in size, ranging from 150 observations to nearly 13,000 observations. They also vary in the number of variables relative to the number of observations and the incidence of the target. The Iris dataset is balanced, compared to an overall outcome rate of 30.9% in the MIMIC II data and just 2.3% in the landslide data. In some cases, these differences impact the way that the UFA system is used. For example, in the case of landslides, the small number of positive instances led us to focus on variables that increase the relative risk, and we found that N-UFA could successfully identify days with a high likelihood of landslide even when the training data were unbalanced.

One possible limitation to the UFA system is that it conducts $nk$ statistical tests in the training phase in order to identify the optimal thresholds, where $n$ is the number of variables and $k$ is the number of potential thresholds. As is well documented, multiple hypothesis testing can inflate the type I error rate and lead to signficant results, even when none exist [2, 5]. This drawback is also present in related methods, such as the minimum p-value approach to finding optimal cut points, and a variety of solutions have been suggested. In this paper, we address the issue through validating the thresholds on previously unseen data. Another possibility is to adjust the p-values in the training phase for multiple testing, using an approach such as the well-known Bonferroni method [5], though that approach was not explored here.

## Conclusion

This paper presents a simple algorithm for identifying univariate thresholds in data. UFA builds on previous work in this area by only considering a subset of the input space, while simultaneously being fully automated.

The thresholds generated by UFA can easily be combined to predict outcomes for previously unseen cases. Though a variety of methods for combining the thresholds exist, in this paper, we introduce N-UFA which classifies observations based on their number of high risk and number of low risk flags. N-UFA greatly reduces the dimensionality of the problem and has similar or better performance than many other commonly used classification techniques, including random forest and logistic regression. In addition to strong predictive performance, this paper highlights several other key advantages of the UFA system:

1. Fully automated; can be used with little a priori knowledge of the data
2. Scales to a large number of variables, even if the number of variables exceeds the number of observations
3. Provides the user with simple rules characterizing relationship between individual variables and the outcome
4. Stable against noise and missing data
5. Useful when the incidence of the target is low
6. Displays results in two dimensions making it easy to interpret and visualize

Future work should focus on better capturing the uncertainty inherent to the thresholds, potentially through methods such as bootstrapping, and formalizing the conditions under which the UFA system performs well.

# Supplemental Materials

**Table A1: List of thresholds for *Iris* dataset, selected based on maximum absolute z-statistic**

| Variable | Threshold | | N | % Versi. | ZStat | ZStat.Abs | Sig |
|---|---|---|---|---|---|---|---|
| Sepal.Length | Less Than | 5.7 | 24 | 87.5% | 3.4 | 3.4 | 1 |
| Sepal.Width | Less Than | 2.4 | 10 | 90.0% | 2.3 | 2.3 | 0 |
| Petal.Length | Less Than | 4.7 | 45 | 97.8% | 5.2 | 5.2 | 1 |
| Petal.Width | Less Than | 1.5 | 48 | 93.8% | 4.4 | 4.4 | 1 |
| Sepal.Length | More Than | 7.0 | 12 | 0.0% | -3.0 | 3.0 | 1 |
| Sepal.Width | More Than | 3.2 | 10 | 20.0% | -1.7 | 1.7 | 0 |
| Petal.Length | More Than | 5.0 | 42 | 2.4% | -5.1 | 5.1 | 1 |
| Petal.Width | More Than | 1.7 | 46 | 2.2% | -5.9 | 5.9 | 1 |

**Figure A1: Automated thresholds for *Iris versicolor* and *Iris virginica*.** *Iris versicolor* is denoted in green and *Iris virginica* is denoted in blue.

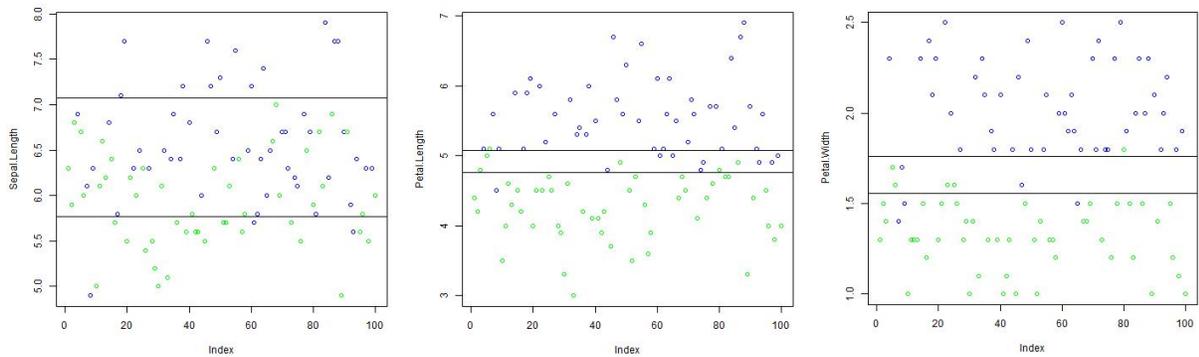